\documentclass[11pt]{article}
\usepackage{acl2021}
\usepackage{times}
\usepackage{latexsym}

\usepackage{graphicx}
\usepackage{amsmath,systeme}
\usepackage{bm}
\usepackage{amsmath}
\usepackage{amsfonts}

\usepackage{algorithm}
\usepackage[noend]{algpseudocode}
\usepackage[export]{adjustbox}
\usepackage{array}
\usepackage{microtype}

\aclfinalcopy 

\title{ABB-BERT: A BERT model for disambiguating abbreviations and contractions}
\author{Prateek Kacker${^1}$\thanks{\hspace*{.5em} Correspondence to: prateek.kacker@novartis.com},  Andi Cupallari${^1}$, Aswin Gridhar Subramanian${^2}$\thanks{\hspace*{.5em} Part of work was done during employment at Novartis} \and Nimit Jain${^1}$ \\
	  Novartis Pharmaceuticals, NJ, USA${^1}$ \\ School of Informatics, University of Edinburgh${^2}$ }

\date{}

\begin{document}
\maketitle

\begin{abstract}
  Abbreviations and contractions are commonly found in text across different domains. For example, doctors' notes contain many contractions that can be personalized based on their choices. Existing spelling correction models are not suitable to handle expansions because of many reductions of characters in words. In this work, we propose ABB-BERT, a BERT-based model, which deals with an ambiguous language containing abbreviations and contractions. ABB-BERT can rank them from thousands of options and is designed for scale. It is trained on Wikipedia text, and the algorithm allows it to be fine-tuned with little compute to get better performance for a domain or person. We are publicly releasing the training dataset for abbreviations and contractions derived from Wikipedia.
\end{abstract}
\section{Introduction}
We use abbreviations and contractions (called "short forms" in this paper) while quickly typing on digital apps. They are used to save time or effort in typing and may be unique to us; therefore, sometimes, only we can understand them. There is no reliable dictionary of short forms to be referred because it can be specific to a context or a person. The short forms often have multiple meanings depending on the domain or the person. In Table  \ref{fig:first_page_table}, consider the sentence \textit{"The doctor saw an AS cd at tl"} written in a notepad by sales representative at pharmaceutical company or local news reporter in Las Vegas. It may be a shorthand for \textit{"The doctor saw an Ankylosing Spondylitis candidate at trial"} for the sales representative or \textit{"The doctor saw an Ace of Spade card at the table"} for the news reporter. Applying downstream NLP AI Algorithms to this shorthand text gives poor performance because they have not been trained on personalized shorthand text. To build better AI systems, we should expand the short forms in the sentences for the domain or the user before using it downstream. Since there is no correct answer for expansions and numerous right choices based on the domain and the user, ranking is the better way to handle short forms text in real-world AI applications.

\begin{table} [t!]
\small
\setlength{\extrarowheight}{1pt}
\begin{tabular}{ |p{2cm}||p{5cm}| }
	\hline
	\multicolumn{2}{|c|}{\textbf{Text notes}} \\
	
	\hline
	\textbf{Notes 1}  & The doctor saw AS cd at tl\\
	\textbf{Notes 2}  & The doctor saw AS cd at tl\\
	\hline
	\multicolumn{2}{|c|}{\textbf{Ground Truth}} \\
	\hline
	\textbf{Notes 1}  & The doctor saw Ankylosing Spondylitis candidate at trial\\
	\textbf{Notes 2}  & The doctor saw Ace of Spades card at the table\\
	\hline
	\multicolumn{2}{|c|}{\textbf{ABB-BERT input}} \\
	\hline
	\textbf{Notes 1}  & The doctor saw at \textit{[ABB]} \textit{[ABB]} at \textit{[ABB]}\\
	\textbf{Notes 2}  & The doctor saw at \textit{[ABB]} \textit{[ABB]} at \textit{[ABB]}\\
	\hline
	\multicolumn{2}{|c|}{\textbf{ABB-BERT outputs (sorted list on rank)}} \\
	\hline
	\textbf{Notes 1}  & \textit{[ABB]}- [Ankylosing Spondylitis, ...]\\
	 & \textit{[ABB]}- [candidate, ...] \\
	 & \textit{[ABB]}-[trial,...] \\
	\textbf{Notes 2}  & \textit{[ABB]}-[Ace of Spades,...] \\
	 & \textit{[ABB]}-[card,...]\\
	 & \textit{[ABB]}-[table,...]\\
	\hline
\end{tabular}

	\caption{Text notes made by one can be ambiguous for others. \textbf{Notes 1} was written by pharmaceutical sales, and \textbf{Notes 2} was written by local news in Las Vegas. \textbf{ABB-BERT} can suggest the best replacement using a fine-tuned model for a domain}
	\label{fig:first_page_table}
\end{table}

The definition of abbreviation is simple, and everyone understands it. For example, \textit{USA} stands for the United States of America, or \textit{MS} stands for Multiple Sclerosis. On the other hand, a contraction is a misspelling or shortening of any word, such as \textit{dr, drs, dctr} etc., for a doctor or \textit{ ptnt, pnt, pt} etc., for a patient. Current spelling correction models fail to capture the correct form for all possible scenarios because of the many reductions of characters in short forms. 

Large NLU language models like BERT \cite{devlin2019bert}, RoBERTa \cite{liu2019roberta}, ALBERT \cite{Lan2020ALBERT:},  or DeBERTa \cite{he2020deberta} are trained on normalized data from different domains but not with personalized or domain-specific short forms and hence reduce the model performance in downstream NLP tasks. For example, in a classification task, the contractions or abbreviation might be critical in determining the class and can lead to a wrong classification (false positive or false negative). To solve this problem, ABB-BERT can normalize the text by ranking the options to find the best choice for abbreviation or contraction, leading to better downstream performance.
 
In the past, much work has been done on normalizing text data. Misspellings (simple spelling mistakes) have been handled well by AI models. Recent work by \citet{tan-etal-2020-tnt} introduced TNT, a model that was developed to learn language representation by training transformers to reconstruct text from operation types typically seen in text manipulation, which they show is a potential approach to misspelling correction. Another AI algorithm Neuspell \cite{jayanthi2020neuspell} is a spelling correction toolkit that captures the context around the misspelled words. We have noticed that misspelling AI models do not perform well with contractions because of loss in information due to contraction and the number of possible variations for the right choice based on domain.
 
 \citet{kreuzthaler-etal-2016-unsupervised} introduced a data-driven statistical approach and a dictionary-based method for the task of abbreviation detection. They show some success of these approaches; however, as their approach depends on a dictionary with a limited number of entries, it cannot be scaled or extended to other domains. \citet{JOOPUDI201871} trained Convolutional Neural Network (CNN) models to disambiguate abbreviation senses and found a 1–4 percentage points higher performance for CNN compared to Support Vector Machines. These results were robust across different datasets.

  \citet{Li2019ANT} showed that topic modeling combined with attention networks could help get better results on abbreviation disambiguation because topics provide the context for the neural networks. To improve the performance on bio-medical data, \citet{Jin2019DeepCB} utilized pre-trained model BioELMO \cite{jin-etal-2019-probing} which gets better-contextualized features of words. Then the features are fed into abbreviation-specific bidirectional LSTMs where the hidden states of the ambiguous abbreviations are used to assign the exact definitions. Recently, \citet{pan2021bertbased} proved that BERT-based algorithms combined with training strategies like dynamic negative sample selection and adversarial training are very effective in Scientific AI domain acronym disambiguation datasets (SciAD) \cite{Veyseh2020}.

The contribution of this paper is threefold.
First, we propose ABB-BERT which uses a ranking algorithm by combining BERT \cite{devlin-etal-2019-bert} and architecture by LaBSE in \citet{labse_feng2020} on short forms options based on context. We introduce a new token \textit{[ABB]} that replaces all short forms. 
Second, we show that ABB-BERT is a practical and scalable way to deal with un-normalized text across domains.
Third, we publicly release the dataset and code \footnote{Dataset and code available at https://github.com/prateek-kacker/ABB-BERT} for ABB-BERT for future work.

\begin{figure*}[t!] 
	\center
	\includegraphics[width=1\textwidth]{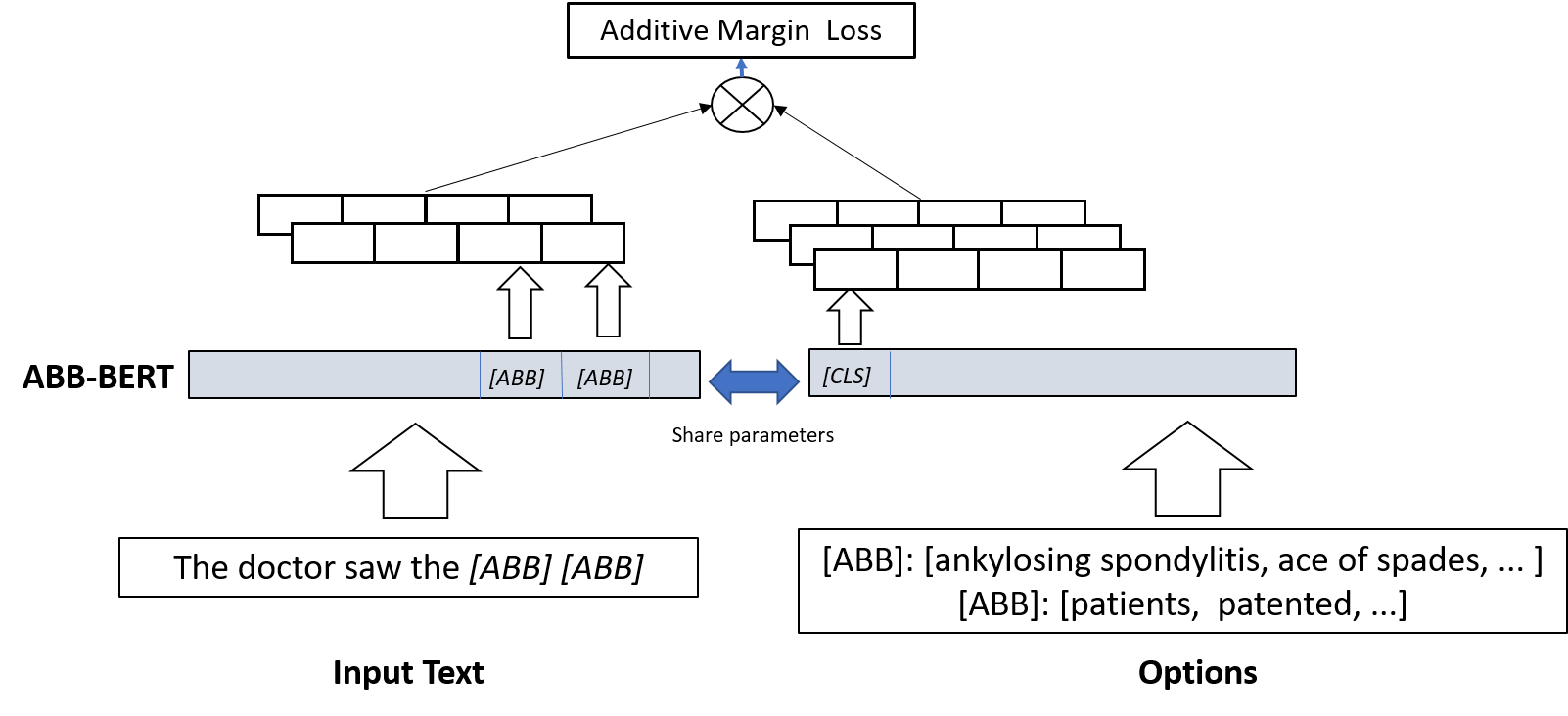}
	\caption{Graphical representation of the training and inference on ABB-BERT. The sentence written by the sales rep is \textit{"The doctor saw the AS ptnt"} but for training, the sentence is modified to \textit{"The doctor saw the [ABB] [ABB]"}. The model is trained to minimize the additive softmax loss of \textit{[ABB]} corresponding to \textit{AS} and \textit{ptnts}. The ground truth for the above example is \textit{"The doctor saw the Ankylosing Spondylitis patient"}. During inference, ABB-BERT ranks the several options given per \textit{[ABB]}. ABB-BERT can be fine-tuned to improve the performance of a domain.}	
	
	\label{fig:dual_encoder}
\end{figure*}

\begin{algorithm}[t!]
	\label{algo:contraction}
	\caption{\textit{contraction}}
	\textbf{Input}: word \\
	\textbf{Output}:List of possible contractions 
	\begin{algorithmic}[1]
		\State Remove any other characters except $a-z$,$A-Z$ and lower case the word
		\State Remove all the vowels $a,e,i,o,u$ 
		\State Select all characters except ${1^{st}}$ character
		\State Find all possible combinations of selected characters without changing order 
		\State Append the first character to each item in the list
		\State Return list
	\end{algorithmic}
\end{algorithm}

\begin{algorithm}[t!]
	\label{algo:abbreviation}
	\caption{\textit{abbreviation}}
	\textbf{Input}:sentence \\
	\textbf{Output}:List of tuples (Abbreviations,expansions) 
	\begin{algorithmic}[1]
		\State Identify capitalized word in sentence and their location
		\State Identify capitalized word sequences with length two or more
		\State If two sequences are seperated by prepositions or conjuctions then connect them to form a sequence
		\State Create a list of tuples (initals of uppercase words in sequence, sequence)
		\State Return list
	\end{algorithmic}
\end{algorithm}
\begin{table} [t!]
	\small
	\centering
	\setlength{\extrarowheight}{1pt}
	\begin{tabular}{ |p{1cm}||p{4cm}||p{1cm}| }

		\hline
		\textbf{Key}  & \textbf{Value} & \textbf{Number of choices} \\
		\hline
		\textbf{ptnt}  & patient, patent, potent, potential ... & 4736 \\
		\hline 
		\textbf{dctr}  & doctor, director, documentary, declaratory ... & 2555 \\
		\hline 
		\textbf{tl}  & table, trial, tool, tuberculosis ... & 81236 \\
		\hline 
	\end{tabular}	
	\caption{Selected examples of key-value pairs in $dict\_cont$. }
	\label{table:dict_cont}
\end{table}

\begin{table} [t!]
	\small
	\centering
	\setlength{\extrarowheight}{1pt}
	\begin{tabular}{ |p{1cm}||p{4cm}||p{1cm}| }

		\hline
		\textbf{Key}  & \textbf{Value} & \textbf{Number of choices} \\
		\hline
		\textbf{as}  & Ankylosing Spondylitis, Ace of Spades, Astronomy and Space ... & 89119 \\
		\hline 
		\textbf{acl}  & Association for Computational Linguistics, Avant Co. Ltd., Albany Club in London ... & 2259 \\
		\hline 
		\textbf{usa}  & United States of America, Urban Songwriter Award, Ultimate Sports Adventure ... & 1608 \\
		\hline 
	\end{tabular}	
	\caption{Selected examples of key-value pairs in $dict\_abb$. }
	\label{table:dict_abb}
\end{table}

\begin{table} [t!]
	\small
	\centering
	\setlength{\extrarowheight}{1pt}
	\begin{tabular}{ |p{3cm}||p{3cm}| }

		\hline
		\textbf{Original Sentence}  & \textbf{With contractions and abbreviations}  \\
		\hline
		\textbf{When I} got to the house, \textbf{Mrs. Everett}, the \textbf{housekeeper}, told me that Hermione was in her room, watching her maid pack.
		& \textbf{WI} got to the house, \textbf{ME}, the \textbf{hs} told me that Hermione was in her room, watching her maid pack. \\
		\hline 
		The Sydney area has been \textbf{inhabited} by \textbf{indigenous} Australians for at least 30,000 years. & The Sydney area has been \textbf{id} by \textbf{ig} Australians for at least 30,000 years  \\
		\hline 
		Bosnian \textbf{claims} of Serbian annexation attempts in 1993 were brought to the \textbf{World Court}. & Bosnian \textbf{cs} of Serbian annexation attempts in 1993 were brought to the \textbf{wc}.\\
		\hline 
		
	\end{tabular}	
	\caption{Selected examples of GLUE Benchmark datasets. They have been manually edited to create training data for ABB-BERT}
	\label{table:glue_examples}
\end{table}

\section{ABB-BERT}
The goal of the algorithm is to rank the options for short forms. As shown in \ref{fig:dual_encoder}, the input sentence $X$ may contain one or more short forms. We assume that short forms' location in the sentence and the character composition is known for training purpose. A typical example of sentence with short form can be seen in Table \ref{table:glue_examples}. We substitute the short forms with $[ABB]$ and the algorithm uses character composition to get several options for short forms, create embeddings  and calculate scores to rank each option.

\subsection{Lookup Tables for Options}
ABB-BERT ranks options based on thousands of choices for expansions for short forms present in the lookup tables \textit{dict\_cont} and \textit{dict\_abb}. To create these tables,  English Wikipedia has been parsed for words for contractions and full forms for abbreviations using the rule-based methods in Algorithm 1 and Algorithm 2, respectively. After observing thousands of short forms used in real-world datasets, these rules were created, which we cannot share publicly. Using these rules, we created key-value pairs for lookup tables, \textit{dict\_cont} and  \textit{dict\_abb}, from words and abbreviations extracted from English Wikipedia. Given an abbreviation or contraction, these lookup tables list words that can be the possible expansion. We can see the output of Table \ref{table:dict_cont} and Table \ref{table:dict_abb}, and this list can be huge; hence scalability is crucial for ABB-BERT. 
\subsection{Model}
ABB-BERT is based on the BERT architecture. In order to make it lightweight, it is pre-trained on an uncased  BERT base model. ABB-BERT requires that every contraction and abbreviation be replaced with $\textit{[ABB]}$ token. Since \textit{[ABB]} is not present in the default vocabulary of BERT, the vocabulary has to be modified to include this special token. Consider a sentence $X$ in dataset $D$.  After the tokenization of $X$,  a sequence of tokens ${(x_1,x_2,...x_n)}$ is generated. In this setup, $x_1$ is the \textit{[CLS]} token for every  sentence. 
 We have already replaced the short forms with the \textit{[ABB]} tokens hence we know their exact locations. For simplicity, let $(x_a,x_b,...)$ represent the \textit{[ABB]} token corresponding to indices $ I = (a,b ...) $ . The  BERT output $\bm{z_1},\bm{z_2},..., \bm{z_n}$ corresponding to each token $x_1,x_2,.., x_n$ can be represented as \begin{equation}
 \bm{z_1},\bm{z_2},..., \bm{z_n} = BERT(x_1, x_2,..., x_n) 
\end{equation} 

On top of the BERT model, there is a feed-forward neural network $f(.)$. The output vectors from BERT,  $\bm{z_i}$, go through this neural network such that $\bm{y_i} = f(\bm{z_i})$. The final combined representation of the output is 
\begin{equation}\label{eq:b}
\bm{y_1},\bm{y_2},...,\bm{y_n} = \bm{ABB\_BERT}(x_1,x_2,...,x_n)
\end{equation}
$\bm{y_1}$ is the corresponding output for always the token representing from \textit{[CLS]} and  $\bm{y_a},\bm{y_b} ... $ for \textit{[ABB]} because of the indices $I$. 

We do similar exercise for short forms. Each short form at  \textit{[ABB]} can have thousands of options and can be found from \textit{dict\_cont} and \textit{dict\_abb} tables. For location \textit{a}, the options are denoted by $\bm{S_a}$ which is list of options $[S_a^1,S_a^2,...,S_a^{o_a}]$ and length of the list is denoted by $o_a$. The tokenizer converts $S_a^1$, the first option to $(s_a^{1,1},s_a^{1,2}, ..., s_a^{1,n})$ and similarly for other options $S_a^2,...,S_a^{o_a}$. Similar to $X$, every option $S_a$ is propogated through $\bm{ABB\_BERT}$. The output is represented for $S_a^1$ is represented as: 
\begin{equation}\label{eq:c}
(\bm{t_a^{1,1}},\bm{t_a^{1,2}},...,\bm{t_a^{1,n}}) 
=\bm{ABB\_BERT}(s_a^{1,1},s_a^{1,2}, ..., s_a^{1,n})
\end{equation} 

The equation \ref{eq:c} is applied to other options in $\bm{S_a}$ also. For options, there will not be any \textit{[ABB]}. \textit{[CLS]} is the first and the only relevant token hence the notations can be simplified by dropping the location information. For instance, ${s_a^{1,1}}$ to ${s_a^{1}}$, ${s_a^{2,1}}$ to ${s_a^{2}}$  etc and similarly for  $\bm{t_a^{1,1}}$ to $\bm{t_a^{1}}$,  $\bm{t_a^{2,1}}$ to $\bm{t_a^{2}}$ etc. The algorithm uses additive margin softmax loss, discussed in the next section, to rank the outcomes.


\subsection{Dual Encoder with Additive Margin Softmax Loss}

The architecture of ABB-BERT with additive margin softmax loss is shown in figure \ref{fig:dual_encoder}. The architecture is similar to the one used by \citet{labse_feng2020}. We use dual-encoders which feeds a scoring function and determines the rank of the alternatives based on the cosine similarity measure, and hence such models are well suited for ranking problems. 
We use the additive margin softmax loss function introduced in \citet{additive_ms_intro}.
Later on, \citet{yang2019improving} used a slightly modified version of this loss, and \citet{labse_feng2020} used it in their language-agnostic LABSE model.

\begin{table*} [ht!]
	\centering
	\setlength{\extrarowheight}{2pt}
	\begin{tabular}{ |p{1.1 cm}||p{4 cm}| p{1.55cm}| p{1.55cm}|| p{1.55cm}|}
		\hline
		& {Metrics} & \textbf{Results A}  &\textbf{Results B} &\textbf{Results C}\\
		\hline
		COLA & Matthews corr.	& 52.6 & 22.8 & \textbf{48.1} \\
		\hline
		SST2&acc	& 93.6 & 20.4 & \textbf{92.9} \\
		\hline

		STS-B&Pearson/ Spearman corr	& 84.9/83.4 & 62.5/61.2 & \textbf{75.0/73.8} \\
		\hline
		QQP&acc./F1		& 71.6/89.2 & 54.5/84.9 & \textbf{65.4/88.3} \\
		\hline
		MNLI Matched&acc.	& 84.5 & 71.3 & \textbf{77.9} \\
		\hline
		MNLI Mismatched&acc.	&83.4 & 72.0 & \textbf{77.1} \\
		\hline
		MRPC&acc./F1	& 86.6/81.6 & \textbf{79.8/75.4} & 75.9/71.5 \\
	    \hline
		QNLI&acc.		& 90.9 & \textbf{83.4} & 82.6 \\
		\hline
		RTE&acc.	        & 64.4 & 56.7  & \textbf{57.8} \\
		\hline
		WNLI&acc.	& 57.5 & \textbf{61.0} & 58.2\\
		\hline

	\end{tabular}
	\caption{ Results of inference of downstream task trained on a single BERT-base-uncased model on GLUE Dataset on the respective tasks. Results A are obtained on the original test datasets. Results B are obtained on test datasets manually edited by introducing short forms. Results C are obtained on the test datasets improved by ABB-BERT by selecting $1^{st}$ option
	}
	\label{table:glue}	
\end{table*}


For this paper, the short forms disambiguation problem is modeled as a ranking problem to find the best option $S_a$ for short form at index $a$ in sentence $X$ where $S_a$ is one of the alternatives in $[S_a^1,S_a^2,...,S_a^{o_a}]$. The ranking of the options is evaluated by the cosine similarity score $\phi$ . For ABB-BERT, $\phi$ scores for all the options at location a, is calculated by calculating cosine similarity $\phi$ between $\bm{y_a}$ and $\bm{t_a^{1}},\bm{t_a^{2}},...,\bm{t_a^{o_a}}$ for options $S_a^1,S_a^2,...,S_a^{o_a}$. Ranking of the options at location $a$ is done by sorting $\phi$ scores.

To train the algorithm, we find the conditional probability $P(S| X)$ for options and for all locations. For example, $S_a^1$, the first option at location $a$ will have $P(S_a^1| X)$ as:
\begin{equation}
P(S_a^1|X) = \frac{e^{\phi(t_a^{1}, y_a)}}{\sum_{i = 1}^{o_a}e^{\phi(t_a^{i}, y_a)}}
\end{equation}

This can be extrapolated to other options and other locations. For training purposes, for each location, the first option is the ground truth option.

Additive margin softmax extends the cosine similarity $\phi$ by introducing a large margin, $m$, only around correct option. The margin improves the separation between the correct option and other options. Moreover, we scale the cosine values using a hyper-parameter $s$ in the equation \ref{eq:most_important}. We select a large value, which accelerates and stabilizes the optimization (see \cite{additive_ms_intro}). Equation \ref{eq:most_important} represents the loss function and is optimized during training.

Substituting for the new scoring functions, the objective loss function for single sentence $X$ becomes:
\begin{equation} \label{eq:most_important}
\mathcal{L} = -\frac{1}{N}\sum_{i=a,b,..}^{I}\sum_{o=1}^{n_i}\frac{e^{s(\phi(t_i^{o}, y_i)-m)}}{\sum_{k = 1}^{n_i}e^{s(\phi(t_i^{k}, y_i)-m)}}
\end{equation}
where 
\[   
m = 
\begin{cases}
\text{1 $\ge$ \textit{m} $\ge$ -1} &\quad\text{\textit{o}=1 or \textit{k}=1}\\
\text{0} &\quad\text{otherwise} \\
\end{cases}
\]

$$ s \gg 1 $$

\subsection{Scalable and personalized ABB-BERT}
ABB-BERT might have to work real-time during inference in some applications to generate options for downstream tasks, though forward pass through BERT over thousands of alternatives can be expensive and time-consuming. Once ABB-BERT gets deployed, it is expected to get better in ranking for a domain, a user, or group of users with new annotations and training runs; hence there should be a personalization phase equivalent to finetuning the model for a person or a domain. In the \textit{personalization phase}, it is not computationally possible to perform a forward pass on an entire list of options or retrain ABB-BERT again as the inference may be on a device with limited compute.
\begin{table} [t!]
	\setlength{\extrarowheight}{2pt}
	\begin{tabular}{ |p{1.5cm}||p{2.5cm}| p{2.5cm}|}
		\hline
		& \textbf{\% Correct outcomes (longest contr.)}  &\textbf{\% Correct outcomes (short contr.)} \\
		\hline
		Wikipedia  & 63.7 & 11.2 \\
		\hline
		Covid Dataset  & 61.7 & 11.0 \\
		\hline
		Apps Review  & 54.4 & 0.0 \\
		\hline
		US Bill  & 58.9 & 0.67 \\
		\hline
		ECTHR  & 70.0 & 13.9 \\
		\hline
	\end{tabular}
	\caption{ Neuspell results on test sets of different domains on contractions. The model performs well with the longest contraction because it has the most number of words. The performance of Neuspell on abbreviations was close to 0 for all datasets.
	}
	\label{table:baseline}
\end{table}
To prepare for the \textit{personalization phase} and to reduce the inference time, \textit{dict\_cont} and \textit{dict\_abb} is parsed for expansions for all possible options and ABB-BERT embeddings $\bm{t_{i}}$ are stored in a lookup table \textit{dict\_embed} for each expansion. The parameters for ABB-BERT and the table \textit{dict\_embed} are then frozen. ABB-BERT is modified by adding a single feed forward layer $g(.)$ parametrized by $\theta$ such that for embeddings $y_i$ for sentences from equation \ref{eq:b} and  embeddings $\bm{t_x^o}$ for options from equation \ref{eq:c}, is modified. Training is done only on layer $g(.)$ to reduce training time.
\begin{equation*}
\bm{u_i^o}=g(\bm{t_i^o},\theta)
\end{equation*}
and $ i \in I$ and for input sentences $\bm{y_i}$ \begin{equation*}
\bm{z_i}=g(\bm{y_i},\theta) 
\end{equation*} 
Here $\bm{u_i^o}$ and $\bm{z_i}$ replaces $\bm{t_i^o}$ and $\bm{y_i}$ respectively in equation \ref{eq:most_important}.   The model parameters are initialized by $\theta_0$ such that $x=g(x,\theta_0)$. During the personalization phase, only parameters $\theta$ are trained which is not computationally expensive and can be done on the device. During inference, ABB-BERT does a forward pass only for input sentence $\bm{y_i}$. ABB-BERT does not need to do a forward pass on options and it can get embeddings $\bm{t_i^o}$ directly from \textit{dict\_embed} and a forward pass with parameters $\theta$.

\section{Experimental Setup}
\subsection{Data Preparation and Pre-training ABB-BERT}
\label{section:datapreparation}
Training of ABB-BERT requires significant preparation of train, test, and validation data. We have taken English Wikipedia and extracted random sentences for datasets. We know that Wikipedia does not have contractions as it is a very clean dataset. Hence we had to create the datasets manually for ABB-BERT based on short forms in algorithm 1 and algorithm 2 respectively. For contractions, 15\% of words in a sentence are selected at random. Using algorithm 1, a random contraction is selected to get options from  \textit{dict\_cont}. Train, test and validation datasets contains $>$1M, $>$100K , $>$100k \textit{[ABB]} tokens respectively.  The ground truth, which is the correct expansion, is always the first word of the options in training, test, and validation datasets. In all the datasets, we have only 50 options per \textit{[ABB]}. In real-world scenarios, there will be thousands of options. Pre-training of ABB-BERT was done on NVIDIA K80 GPUs for a week on Wikipedia training data with Adam optimizer and a $lr$ equal to $5e-06$. After hyper parameter optimization, $m$ was chosen to be 0.8, and $s$ was 30.

\subsection{Results}
Each sentence in ABB-BERT can have multiple \textit{[ABB]} and performance is calculated at each location at $I = (a,b,...)$ 
There are two metrics relevant to this experiment. First is the average of rank (R) of the correct ground truth option, and second is average of difference ($Dif$) between cosine value ($\phi$) of input sentence \textit{[ABB]} \& correct option which is the first option in training data and average cosine value of the input sentence $[ABB]$ \& rest of the options

\begin{equation} \label{eq:d}
Dif = \phi(\bm{t_a^{1}},\bm{y_i}) - (\sum_{n = 2}^{o_a}\phi(\bm{t_a^{n}},\bm{y_i}))/(o_a-1)
\end{equation}

We understand that the best average rank of the model outcome on the test set is 1, and $Dif$ should be close to  $m$ on average. The larger the value of $Dif$, the better ABB-BERT is in predicting the outcome.

\begin{table*} [ht!]
	
	\centering
	\setlength{\extrarowheight}{1pt}
	\begin{tabular}{ |p{1.6cm}|p{1.2cm}|p{1cm}|p{1cm}||p{1.2cm}|p{0.8cm}|p{1.4cm}|p{1.3cm}|p{1.3cm}|p{1.3cm}|}
		\hline
		& \multicolumn{3}{|c|}{\textbf{Pre-personalization }}& \multicolumn{6}{|c|}{\textbf{Post-personalization}} \\
		\hline
		& Avg. Rank over 50 options & [ABB] count & Avg. Diff & \%(Top 3 ranks) in test set & Avg. Rank  & Avg. Rank improvement & count of [ABB] Rank increase & Avg. Rank decrease & count of [ABB] Rank decrease \\
		\hline
		Wikipedia (Pre-training)  & 1.45 & 125079 & 0.67 & - & - & - & -& -& -\\
		\hline
		Covid Dataset & 1.58 & 16218  & 0.61 & 95.7 & 1.57 & 2& 174 & 1.18& 130 \\
		\hline
		Application Review & 1.42 & 9150  & 0.67 & 97.7  & 1.32 & 8.89& 140 & 1.6 & 164\\
		\hline
		US Bill  & 1.51 & 29470  & 0.64 & 96.5 & 1.46 & 5.17& 396 & 1.49 & 332  \\
		\hline
		ECTHR  & 1.26 & 22295  & 0.67 & 98.5 & 1.25 & 3.24& 144 & 1.2& 254\\
		\hline
	\end{tabular}
	\caption{ ABB-BERT performance on different domain data pre and post personalization phase. In pre-personalization phase, ABB-BERT was used without any domain training. The results are for short forms identified together in a sentence. m is 0.8 and Avg Diff is very close to it. The average rank without training is very high leaving little scope of big improvement. However there is improvement seen in rank of the correct option in post-personalization phase	}
	\label{table:ABB-BERT}
\end{table*}
In the first experiment, we evaluate the impact of short forms on any downstream task. In order to model the impact, we took GLUE Benchmark \cite{wang-etal-2018-glue} tasks as a downstream task. Table \ref{table:glue} column $\textbf{Results A}$ show the performance of the BERT-base-uncased model on each task without any changes to test data. We manually introduced short forms in test sets of each task using techniques mentioned in section \ref{section:datapreparation}. There is a marked reduction of performance in most datasets, as shown in table \ref{table:glue} in column $\textbf{Results B}$. Then we corrected each test set with ABB-BERT predictions selecting only the $1^{st}$ rank option from 50 options. The performance of the new test set is shown in the table \ref{table:glue} in column $\textbf{Results C}$. 
In the second experiment, we wanted to measure the model performance improvement after the personalization phase. Hence we tested it out on three domain datasets which were bio-medical, legal, and reviews datasets. For the biomedical domain, the Covid-19 QA dataset by \citet{moller2020covid} was used. For the legal domain, US Legislation Corpus by \citet{kornilova2019billsum} and European Court for Human Rights (ECTHR) database by \citet{chalkidis-etal-2019-neural}  was used. For the technical domain, the Android Applications User Review dataset by \citet{grano2017android} was used. Sentences from paragraphs were extracted for train, validation, and test datasets. The lookup tables \textit{dict\_cont} and \textit{dict\_abb} were used to create short forms for all the datasets and parameters $\theta$ of $g(.)$ were trained keeping $\bm{ABB\_BERT}$ parameters static for this experiment. The number of options for every $[ABB]$ was 50 for every dataset. \\
ABB-BERT performance results on a test set are shown in table \ref{table:ABB-BERT}. Without any training of parameter $\theta$, ABB-BERT does very well in the pre-personalization phase. The performance on the test set in post-personalization gets better though not noticeable because the average rank is close to 1 in pre-personalization phase. \\
In the third experiment, we wanted to compare our work with existing work. We could not find the exact equivalent for this work, but we still decided to baseline this work for contractions using NeuSpell by \citet{jayanthi2020neuspell}. Neuspell is an excellent algorithm for misspellings, but when exposed to contractions, it makes many mistakes. Results of the baseline can be found in table \ref{table:baseline}. As expected, NeuSpell does well for lengthiest contractions than shortest contractions. The performance of Neuspell was close to 0 for all the abbreviation datasets. Hence, the results are not shown in the table \ref{table:baseline}. \\
For abbreviations, we tested out the algorithm on the SciAD dataset from \citet{Veyseh2020}. ABB-BERT, without training, gives an average rank of 1.76, which is lower compared to the best model by \citet{Jin2019DeepCB}. It is because the loss function of the algorithm is designed for ranking on large number of options. However, the performance improves after the personalization phase, with an average ranking of 1.55.

\subsection{Visualizing ABB-BERT results}
\label{appendix:understanding_abb_bert_results}
In the datasets, ABB-BERT is given only 50 options. It does a great job in predictions with more than 90\% performance for the top 3 choices. If we look at the results, we see that most of the time, the one of top choices can make a correct substitution for $[ABB]$ in a sentence. Table \ref{table:good_choice} shows the options that scored high and make much sense. It shows that model can learn grammar and understands language well. However, the model does not consider commonsense or missing context in ranking. Table \ref{table:poor_choice} shows where the model makes mistakes in predictions because of inherent challenges in this task.

\begin{table*} [t!]
	\small
	\setlength{\extrarowheight}{1pt}
	\begin{tabular}{ |p{4cm}||p{4cm}||p{5cm}| }

		\hline
		\textbf{Original Sentence}  & \textbf{With $[ABB]$} & \textbf{top 5 alternatives and cosine scores} \\
		\hline
		Young redheaded man holding two bicycles near beach.  & Young redheaded man holding [ABB] [ABB] near beach.
		& \textbf{ABB1}:(\textbf{two}, 0.99), (twag, 0.20), (twili,0.20), (tmfw,0.20), (townian,0.20) \textbf{ABB2}:(\textbf{bicycles}: 0.86), (berchy: 0.20), (binchy: 0.20), (bakley: 0.20), (besyde: 0.20) \\
		\hline 
		This problem has been \textbf{previously} studied for zero-shot object \textbf{recognition} but there are several key differences. & This problem has been [ABB] studied for zero-shot object [ABB] but there are several key differences. & \textbf{ABB1}:(\textbf{previously}, 0.99), (provincial, 0.98), (privateering, 0.2), (pāval, 0.2), (primavera, 0.2) \textbf{ABB2}: (\textbf{recognition}, 0.99), (recréation, 0.26), (retroactive, 0.21), (rectification, 0.21), (revoluction, 0.20), \\
		
		\hline 
		a \textbf{vivid} \textbf{cinematic} portrait. & a [ABB] [ABB] portrait. & \textbf{ABB1}:(\textbf{vivid}, 0.99), (vmvs, 0.20), (vhvi, 0.20), (vitruvius, 0.20), (vouvantes, 0.20) \textbf{ABB2}: (\textbf{cinematic}, 0.99), (christini, 0.20), (ciston, 0.20), (coeffient, 0.20), (clairant, 0.20) \\

		\hline
	\end{tabular}	
	\caption{Selected examples of GLUE Benchmark datasets. The models made an accurate predictions on the options it was given. The model understands grammar and takes in context in the sentence}
	\label{table:good_choice}
\end{table*}

\begin{table*} [t!]
	\small
	\setlength{\extrarowheight}{1pt}
	\begin{tabular}{ |p{4cm}||p{4cm}||p{7cm}| }

		\hline
		\textbf{Original Sentence}  & \textbf{With $[ABB]$} & \textbf{top 5 alternatives and cosine scores} \\
		\hline
		"It's our judgment that the possible avenues to a peaceful resolution were \textbf{not} fully explored at the Tokyo conference," \textbf{U.S. State Department} spokesman \textbf{Richard Boucher} said.  & " It's our judgment that the possible avenues to a peaceful resolution were [ABB] fully explored at the Tokyo conference," [ABB] spokesman [ABB] said. &  \textbf{ABB1}: (\textbf{not} ,0.99), (nudator, 0.204), (nafat, 0.204),(ndkt, 0.204), (nonotic, 0.203) \textbf{ABB2}:(United States Delegation, 0.99), (Ukrainian Second Division, 0.99),(Ukrainian Soviet Division, 0.99), (Ukrainian Social Democratic, 0.99), (Union of Social Democrats, 0.99), \textbf{ABB3}: (road between,0.99), (Rob Bradley, 0.99), (Rosario Blanco, 0.99), (Ralph Barbara,0.99), (Roger Barclay,0.99) \\
		\hline 
		\textbf{Maude and Dora} saw a \textbf{train} coming & [ABB] saw a [ABB] coming & \textbf{ABB1}:(Mountain Daughter,0.99), (Mo Due, 0.99), (Molino Dam, 0.99), (Mustard Digital, 0.99), (\textbf{Maude and Dora}, 0.99) \textbf{ABB2}:(\textbf{train}, 0.99), (tegne, 0.20), (tanshi, 0.20), (thunderer, 0.20), (trumain, 0.20), (tenelea, 0.20), \\
		\hline 
		\textbf{Alan J. Konigsberg is} related to \textbf{Levy Phillips \& Konigsberg} & [ABB] [ABB] related to [ABB] & \textbf{ABB1}:(All Japan Kick,0.99), (Archbishop John Kemp,0.52), (Arbab Jehangir Khan,0.35), (American John Kendrick,0.3), (Albert James Kingston,0.29) \textbf{ABB2}:(is, 0.99), (istd, 0.20), (isthmo, 0.20), (inscs, 0.20), (istres,0.20), \textbf{ABB3}:(Lord Palatine of Kyiv, 0.99), (Liverpool Park Keepers, 0.99), (La Palabra Kilometros, 0.99), (Long Pine Key, 0.99), (Lalitha Priya Kamalam,0.91) \\
		\hline
	\end{tabular}	
	\caption{Selected examples of GLUE Benchmark datasets. The sentences are hard to predict because of the model outcome might be correct grammatically but does not match the ground truth. Is some cases enough information is not provided as input to make a correct prediction}
	\label{table:poor_choice}
\end{table*}

\section{Conclusions and future work}
In this work, we propose ABB-BERT for abbreviation and contraction disambiguation. ABB-BERT tackles both of these irregularities in the text simultaneously and has the advantage that it takes into account the context in the sentence when ranking the possible alternatives. We designed it on Wikipedia and tested it on domain data also. The model may have a hard time getting the right options if they are grammatical appropriate based on context but maybe wrong by commonsense. Future work can help improve the model and suggest all \textit{[ABB]} at the same time based on commonsense and missing context like geographical location, history of notes etc.


%
%

\bibliographystyle{acl_natbib}
\bibliography{anthology}


\end{document}